\documentclass[letterpaper]{article} 
\usepackage{aaai25}  
\usepackage{times}  
\usepackage{helvet}  
\usepackage{courier}  
\usepackage[hyphens]{url}  
\usepackage{graphicx} 
\usepackage{natbib}  
\usepackage{caption} 
\usepackage{algorithm}
\usepackage{algorithmic}

\usepackage{subcaption}
\usepackage{graphicx}
\usepackage{amsmath}

\usepackage{newfloat}
\usepackage{listings}
\DeclareCaptionStyle{ruled}{labelfont=normalfont,labelsep=colon,strut=off} 
\floatstyle{ruled}
\newfloat{listing}{tb}{lst}{}
\floatname{listing}{Listing}
\nocopyright 

\title{From Dashcam Videos to Driving Simulations: Stress Testing Automated Vehicles against Rare Events}
\author{%
  Yan Miao\textsuperscript{\rm 1}, Georgios Fainekos\textsuperscript{\rm 2}, Bardh Hoxha\textsuperscript{\rm 2}, Hideki Okamoto\textsuperscript{\rm 2}, Danil Prokhorov\textsuperscript{\rm 2}, Sayan Mitra\textsuperscript{\rm 1}
}
\affiliations{
    \textsuperscript{\rm 1} University of Illinois at Urbana-Champaign,
    \textsuperscript{\rm 2} Toyota Research Institude North America
}

\begin{document}

\maketitle

\begin{abstract}
    Testing Automated Driving Systems (ADS) in simulation with realistic driving scenarios is important for verifying their performance.
    However, converting real-world driving videos into simulation scenarios is a significant challenge due to the complexity of interpreting high-dimensional video data and the time-consuming nature of precise manual scenario reconstruction. 
    In this work, we propose a novel framework that automates the conversion of real-world car crash videos into detailed simulation scenarios for ADS testing. 
    Our approach leverages prompt-engineered Video Language Models (VLM) to transform dashcam footage into SCENIC scripts, which define the environment and driving behaviors in the CARLA simulator, enabling the generation of realistic simulation scenarios.
    Importantly, rather than solely aiming for one-to-one scenario reconstruction, our framework focuses on capturing the essential driving behaviors from the original video while offering flexibility in parameters such as weather or road conditions to facilitate search-based testing. 
    Additionally, we introduce a similarity metric that helps iteratively refine the generated scenario through feedback by comparing key features of driving behaviors between the real and simulated videos. 
    Our preliminary results demonstrate substantial time efficiency, finishing the real-to-sim conversion in minutes with full automation and no human intervention, while maintaining high fidelity to the original driving events. 
\end{abstract}

\section{Introduction}
The rapid advancements in Automated Driving Systems (ADS) technology have created an urgent need for robust and realistic testing environments to assure reliability of ADS \citep{7795548}. 
Real-world scenarios, such as crash videos and near-miss events, offer valuable insights into the diverse conditions ADS must navigate, making them an essential source for improving ADS testing.
However, replicating these scenarios in real-world settings is both dangerous and impractical. 
Therefore, converting real-world driving videos into simulation scenarios is a necessary solution, but this process also poses several challenges: the high dimensional video data significantly limits the development of automated methods, while manually reconstructing these scenarios can take experts hours to complete.
This time-consuming process underscores the need for a more efficient and automated approach.

In this paper, we propose a novel framework that automates the conversion of real-world driving videos into detailed simulation scenarios. 
Our approach leverages prompt-engineered Video Language Models(VLMs) to transform dashcam videos into SCENIC scripts, enabling the automatic generation of realistic simulations in CARLA. 
In addition, rather than aiming solely to create an exact digital copy of the dashcam video, our framework prioritizes capturing the most essential driving behaviors while maintaining flexibility in parameters such as weather and road conditions. This flexibility supports search-based testing, a methodology that systematically varies parameters to identify edge cases and vulnerabilities in ADS.
Furthermore, we introduce a similarity metric that iteratively refines the generated simulations by comparing key driving features between the real and simulated videos. 
The contributions of this work are fourfold: (1) an automated video-to-simulation pipeline that removes the need for manual scenario construction, (2) a similarity metric that bridge the gap between real and simulated scenarios, (3) an iterative feedback loop for scenario refinement using neural network-based feedback from VLM to capture core driving behaviors, and (4) a significant improvement in time efficiency, reducing scenario generation from hours to minutes while maintaining high fidelity to the original events. 

\section{Related Work}
Testing ADS in simulation environments has been the subject of extensive research \citep{7795548}. 
Existing autonomous vehicle simulation platforms such as CARLA \citep{DBLP:journals/corr/abs-1711-03938} and LGSVL \citep{9294422} allow researchers to generate and manipulate driving scenarios in controlled environments. 
Efforts such as the SCENIC language \citep{fremont-pldi19} enables search-based testing(SBT) so that the generated scenario can be a seed for search based testing with respect to temporal logic requirements \citep{dreossi2019verifai, TuncaliEtAl2020tiv}, safe driving rules \citep{HekmatnejadHF2020itsc} and traffic laws \citep{SunEtAl2022ase}.
However, accurately designing these scenarios to closely resemble real-world events remains a challenge.

Recently, efforts have shifted toward automatic real-to-simulation (real-to-sim) conversion approaches that use video data to guide scenario generation. 
For instance, Bai et al. 
 \citep{10.1145/3633463} introduced a system that extracts key information from videos to create driving scenarios in simulation environments. 
In \citep{elmaaroufi2024generatingprobabilisticscenarioprograms}, the authors automatically generate driving scenarios from police crash reports.
Similarly, \citep{10378107} focus on learning realistic human behaviors in real-life scenarios and use learned models to improve simulations.
NVIDIA's STRIVE \citep{rempe2022generatingusefulaccidentpronedriving} generates accident-prone driving scenarios by modifying 2D trajectories, but this method is based on controlled scenarios rather than real-world crash videos. 
Another approach, DEEPCRASHTEST \citep{9197053}, converts dashcam footage into crash tests by extracting 3D vehicle trajectories but lacks an iterative refinement process to improve simulation accuracy.

While these approaches represent meaningful progress, existing methods are often limited by a reliance on pre-defined trajectories or fail to incorporate iterative feedback to refine the generated scenarios. 
Furthermore, their primary goal has been to ensure that the generated scenarios are as identical as possible to the original ones (e.g. optimizing on the tracking performance for KITTI tracking dataset \citep{Fritsch2013ITSC}). 
In contrast, we want to focus on capturing the core driving behaviors from the original video while introducing flexibility in other parameters, such as weather and road conditions. 
This flexibility enables search-based testing, allowing for systematic exploration of environmental variations to identify vulnerabilities in ADS and improve their robustness.

To address these gaps, we chose to use VLMs as they enable more flexible and scalable video-to-language translation. Recent works, such as DriveDreamer-2 \citep{zhao2024drivedreamer2llmenhancedworldmodels}, employ LLMs to generate user-defined driving videos by translating queries into agent trajectories and maps for simulation. Similarly, Text-to-Drive \citep{nguyen2024texttodrivediversedrivingbehavior} leverages knowledge-driven language descriptions to synthesize diverse driving behaviors. While these works demonstrate the potential of foundational models in simulation-based driving applications, they either primarily limit to synthetic data and do not leverage real-world driving videos or focus on user-defined queries.

In contrast, our framework integrates real-world driving data, such as dashcam crash videos, to generate simulation scenarios that not only capture critical driving behaviors but also provide flexibility for environmental parameters like road type and weather. Through few-shot prompt engineering, we tailor VLMs to create SCENIC scripts for scenario generation, bridging the gap between exact scenario replication and the need for diversity in simulation environments. This approach ensures realistic and behaviorally accurate test cases while supporting search-based testing for robust ADS validation.

\section{Real-to-Sim Scenario Generation Framework}
\label{sec:framework}
Given a real-life vehicle crash video (e.g., a dash camera recording), our objective is to generate corresponding simulation scenarios that accurately capture the core driving behaviors. In this paper, we define scenario as a driving event that includes environmental conditions (e.g., weather, road types), vehicle dynamics (e.g., speed, acceleration), and driving behaviors between entities (e.g., overtaking behavior or collisions).
This approach enables robust and safe testing of Automated Driving Systems (ADS) by leveraging real-life close-call situations within a controlled simulation environment.
Our framework consists of 4 components: (1) conversion of real-world video into SCENIC scripts, (2) generation of simulation videos from SCENIC scripts, (3) similarity analysis between the real and simulated videos, and (4) iterative refinement to ensure the simulated video's consistency with the original scenario. 

\subsection{Video-to-Text Generation}

\begin{figure*}[ht!]
    \centering
    \begin{subfigure}[b]{0.48\textwidth}
         \centering
         \includegraphics[width=0.8\textwidth]{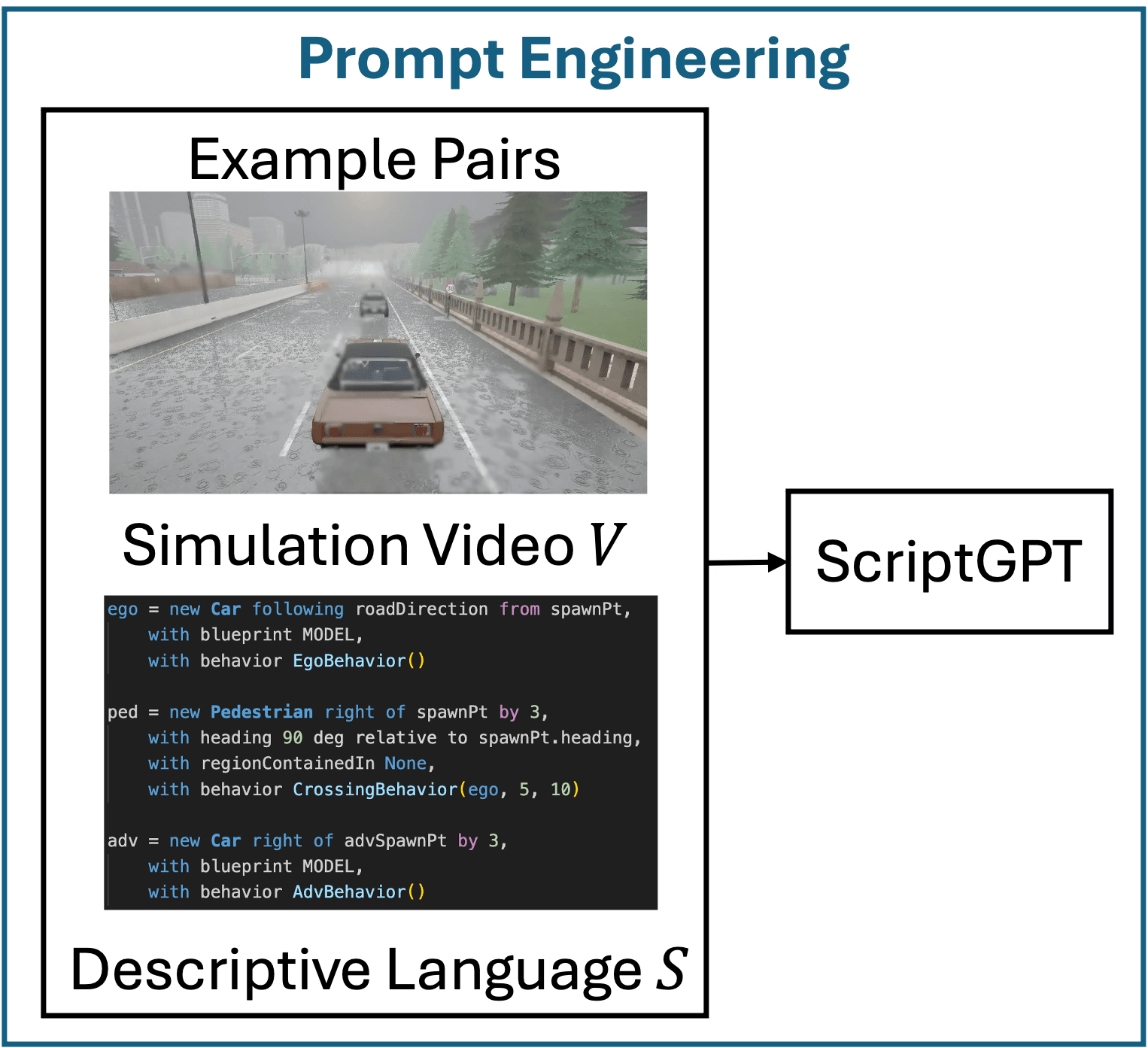}
         \caption{\small \textbf{ScriptGPT}:  A Video Language Model derived from GPT-4o, developed through prompt engineering with paired examples of simulation videos and their corresponding descriptive SCENIC scripts.}
         \label{fig:prompt-scenic}
     \end{subfigure}
     \hfill
     \begin{subfigure}[b]{0.48\textwidth}
         \includegraphics[width=0.8\textwidth]{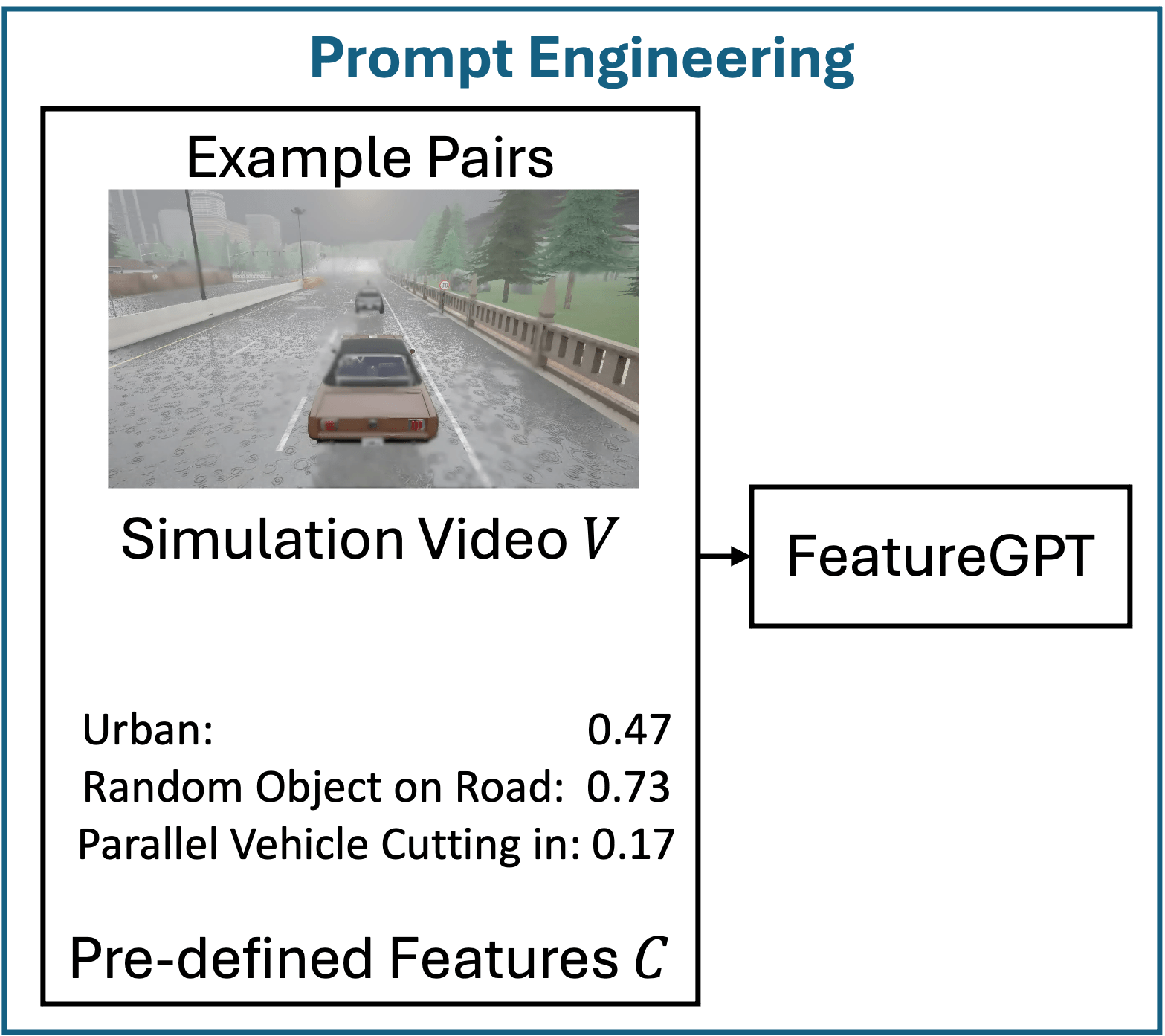}
         \caption{\small \textbf{FeatureGPT}:  A Video Language Model derived from GPT-4o, developed through prompt engineering with paired examples of simulation videos and their corresponding pre-defined features.}
         \label{fig:prompt-feature}
         \end{subfigure}
        \caption{Train {\em ScriptGPT} and {\em FeatureGPT} using Prompt Engineering}
        \label{fig:prompt}
\end{figure*}

First, we convert the input video into a descriptive script called SCENIC.
SCENIC is a probabilistic programming language designed for describing driving scenarios with a high level of precision, enabling detailed specification of environmental factors, road layouts, and vehicle behaviors for use in simulation platforms like CARLA. This conversion is accomplished using a prompt-engineered version of the pre-trained GPT-4o model.

Prompt engineering involves designing and refining input prompts to guide large foundational models, such as GPT-4o \citep{openai2024gpt4technicalreport}, toward producing accurate and desired outputs. 
In this case, prompt engineering is especially effective for generating detailed scenario descriptions (e.g., SCENIC scripts) from real-world driving videos. 
During the prompt engineering process, we improve the pre-trained GPT-4o model by providing it with multiple "positive-example" pairs $(V_i, S_i)$, where $V_i$ is a simulation video generated in CARLA using the corresponding SCENIC script $S_i$ that describes the scenario, as illustrated in Figure \ref{fig:prompt-scenic}. 
Through this process, the new Video-Language Model, referred to as {\em ScriptGPT}, learns to map key visual elements, such as weather, road conditions, and vehicle behaviors, into structured and accurate scenario description languages. 
After sufficient prompt engineering, ScriptGPT is capable of generating a SCENIC script $S_{out}$ for real-world crash video $V_{real}$. 

Although the initial prompt engineering process was conducted using simulation videos paired with their SCENIC scripts, through empirical testing, we found this approach could be extendable to real-world driving videos because the underlying visual and descriptive patterns (e.g., road layouts, traffic behaviors, and environmental factors) are consistent across both domains, allowing the model to effectively generalize its learned capabilities and accurately capture real-world scenarios.

\subsection{Text-to-Video Generation}

\begin{figure*}[htbp]
    \centering
    \includegraphics[width=0.8\linewidth]{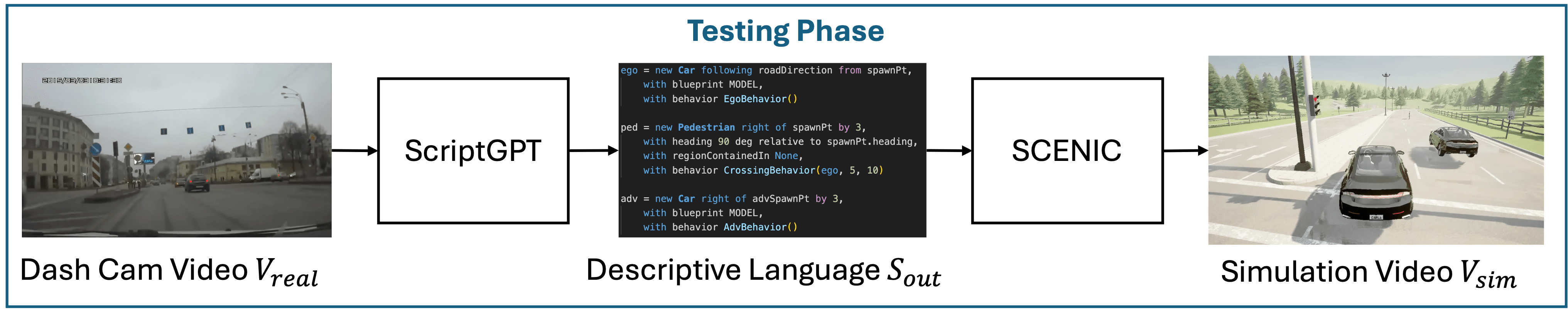}
    \caption{\small After prompt engineering, the dash cam video is fed into {\em ScriptGPT}, which synthesizes descriptive language in the SCENIC format. This SCENIC script can then be executed in CARLA to generate a corresponding testing scenario in simulation.}
    \label{fig:inference-time}
\end{figure*}

We convert the descriptive language $S_{out}$ to simulation video $V_{sim}$ using SCENIC \citep{DBLP:journals/corr/abs-1809-09310, fremont-pldi19}. 
SCENIC is a programming language tool designed for specifying driving scenarios through environmental factors, vehicle behaviors, and road conditions. 
Once we have the SCENIC script $S_{out}$ generated from the real crash video using the prompted-engineered model, we feed it into the SCENIC framework to synthesize a simulation scenario in CARLA, as shown in Figure \ref{fig:inference-time}. 
The script $S_{out}$ serves as the textual representation of the scene, encoding environmental conditions (e.g., weather, traffic), vehicle dynamics, and road types. 
The output is a new simulation video $V_{sim}$, which visually represents the scenario described in $S_{out}$.

\subsection{Similarity Check}

Next, we perform a similarity check on the simulated scenario $V_{sim}$ and the original $V_{real}$, measured on a set of predefined features.
Ideally, the simulated and original scenario should closely match, allowing us to seamlessly replace the ego vehicle with any ADS (e.g. Baidu's Apollo planner \citep{DBLP:journals/corr/abs-1807-08048} and controller) for testing. 
However, due to the complexity of real-world scenarios, even with prompt engineering, discrepancies often arise, where the generated simulation may miss key features or introduce extra ones. 

To address this, we introduce a similarity metric to ensure that the generated video captures the most important features from the original video. 
We predefine a set of crucial feature categories, such as the most critical and frequently encountered driving behaviors and environmental conditions, to examine the original crash video $V_{real}$ with the generated simulation $V_{sim}$. 
Then, we use another prompt-engineered transformer model, {\em FeatureGPT} (depicted in Figure \ref{fig:prompt-feature}), to output a predicted probability (from 0 to 1) for each predefined feature category for a given video. 
Next, the similarity score vector, $Sim(V_{real}, V_{sim})$, is calculated as a vector of differences between the predicted probabilities across the predefined categories: 

\begin{multline*}
Sim(V_{real}, V_{sim}) = [C_{real_1} - C_{sim_1}, C_{real_2} - C_{sim_2}\\
, \dots, C_{real_n} - C_{sim_n}],
\end{multline*}

where $C_{real_i}$ and $C_{sim_i}$ represent the predicted probabilities of $V_{real}$ and $V_{sim}$ for feature $i$. 

\subsection{Iterative Refinement}

\begin{figure*}[htbp]
    \centering
    \includegraphics[width=0.8\linewidth]{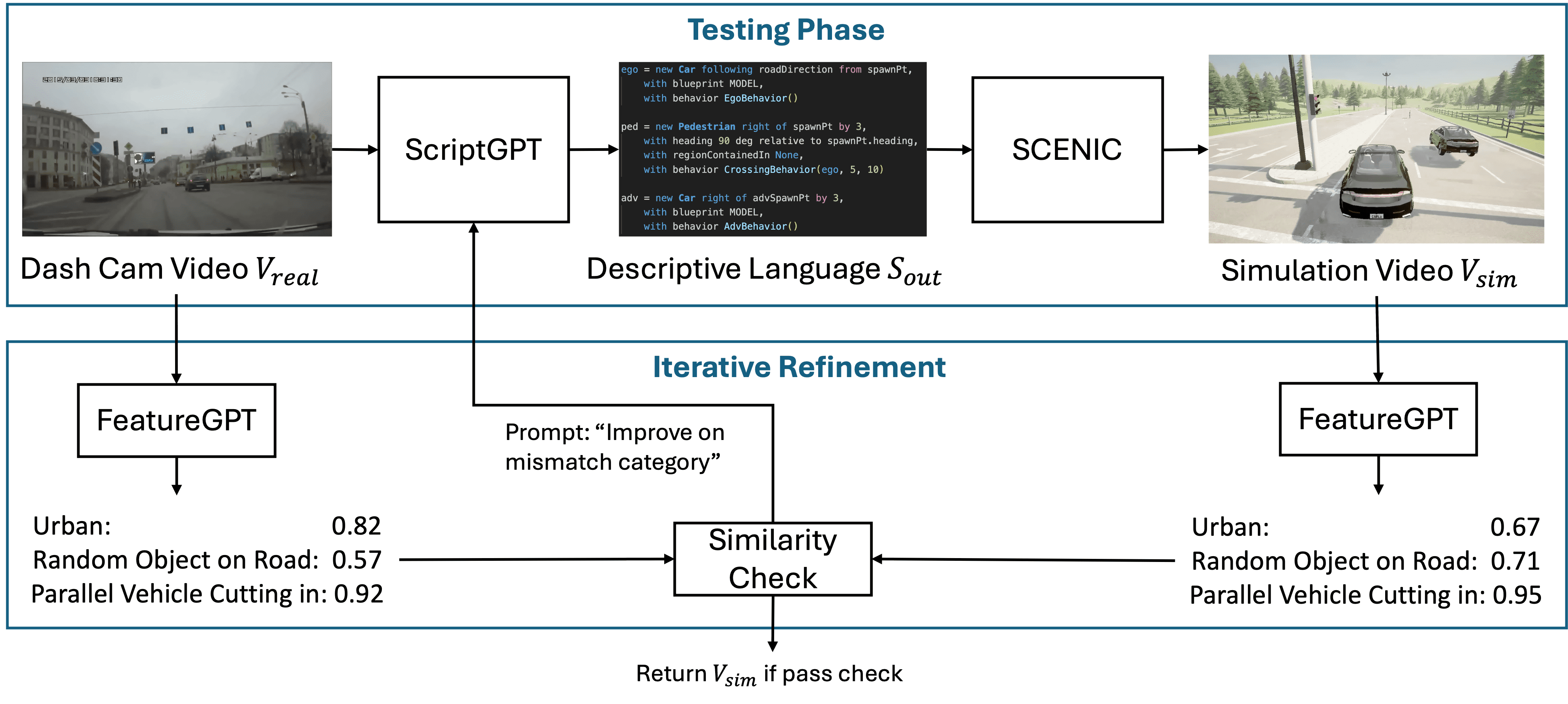}
    \caption{\small \textbf{Iterative Refinement Process}: After obtaining the simulated video $V_{sim}$ from {\em ScriptGPT} and SCENIC, both the original and simulated videos are fed into {\em FeatureGPT} to evaluate the probabilities of predefined features. If the difference of any feature between the original and simulated videos exceeds a certain threshold, we iteratively refine {\em ScriptGPT} by incorporating additional feedback into the SCENIC script, guiding further scenario adjustments until the similarity improves.}
    \label{fig:iterative-refinement}
\end{figure*}

Finally, we perform iterative refinement on the {\em ScriptGPT} with the help of similarity check, as illustrated in Figure \ref{fig:iterative-refinement}.  
If the absolute difference for a given category $i$ is above a predefined threshold $\tau_i$, we refine the SCENIC script $S{out}$ by feeding the discrepancy for that category back into {\em ScriptGPT} as an additional prompt (e.g., "there shouldn't be a leading vehicle overtaking behavior, please improve on that"). 
This feedback allows {\em ScriptGPT} to adjust the SCENIC script $S_{out}$ accordingly, generating a new version of the script and producing a new simulation video $V_{sim'}$.

Although a single SCENIC script can produce multiple simulation videos with consistent core driving behaviors but variations in other parameters (e.g., road types or vehicle configurations), we utilize only a single video for similarity checks and iterative refinement, because multiple simulation videos has the same driving behaviors and refining on a video is sufficient to ensure that the generated scenario aligns with the essential driving behaviors of the original video. 
This approach provides the necessary focus for refining critical features while maintaining flexibility for generating diverse scenarios, enabling search-based testing.

The process is repeated iteratively until the difference for each category falls below the predefined threshold, i.e., $\|Sim(V_{real}, V_{sim'})_i\| \leq \tau_i$. 
Once the similarity across all categories pass the threshold check, the final simulation scenario is ready for testing ADS.

\section{Experiments \& Analysis}
\subsection{System Setup}

\paragraph{Dataset} We obtain the collision videos from the Car Crash Dataset (CCD) \citep{BaoMM2020}. 
CCD is chosen because it contains real traffic accident videos captured by dashcams mounted on driving vehicles, potentially providing a rich source for developing and testing ADS. Moreover, our framework is also relevant for near misses events.

\paragraph{Simulator} We use CARLA~\citep{DBLP:journals/corr/abs-1711-03938}, an open-source platform designed to support the development and validation of autonomous driving systems. 
CARLA is selected for its realistic physics engine and high-fidelity environmental rendering, making it ideal for generating the simulation scenarios needed in our framework.

\paragraph{Description Language SCENIC} We use SCENIC as the description language for driving scenarios due to its similarity to Python, which aligns well with GPT-4o's input data \citep{openai2024gpt4technicalreport}, making prompt engineering doable and more straightforward. 
Furthermore, SCENIC integrates seamlessly with the CARLA simulator and it is highly effective at specifying complex driving scenarios, making it well-suited for our application.

\paragraph{Prompt-Engineered {\em ScriptGPT}}

To construct the {\em ScriptGPT} model, we begin by carefully and manually writing 20 SCENIC scripts that cover diverse driving scenarios, such as overtaking, cruising, sudden stops due to obstacles, and turns in varying road and weather conditions. 
Using the SCENIC library and CARLA, we generate corresponding videos for each scenario and pair them with their respective SCENIC scripts, forming 20 $(V_i, S_i)$ pairs. 
These pairs are then used as input data for prompt engineering GPT-4o, selected for its ability to learn and generalize from a wide range of examples. 

The empirical design choice we made is to design only 20 pairs because through empirical testing, 20 scenarios are sufficient to cover most of the common interesting scenarios. We believe we only need a very small number of prompt engineering due to the powerful generalization ability of pretrained GPT-4o.
Moreover, in practice, since GPT-4o API does not natively accept video formats like .mp4, we preprocess the videos by sampling frames and concatenating them into an n-dimensional array. 
The same preprocessing technique is applied during testing phases to ensure consistency between training and testing phases, and we also apply the same to the {\em FeatureGPT}.

\paragraph{Prompt-Engineered {\em FeatureGPT}}

We use {\em FeatureGPT} to enhance our framework's ability to recognize specific driving behaviors. 
First, we predefine 10 driving feature categories, as shown in Table \ref{table: pre-defined category}. 
Next, we create 20 SCENIC scripts representing scenario videos where these features may or may not appear. 
Each video is paired with a corresponding 10-dimensional feature vector (e.g., [parallel vehicle overtaking: 0, ... , leading vehicle stopped: 1], where 0 indicates absence and 1 indicates presence).
The input data is used to prompt-engineer GPT-4o into {\em FeatureGPT}, which outputs a 10-dimensional probability vector for each video during inference. 
This allows us to compare and categorize driving behaviors between the original video $V_{real}$ and the generated video $V_{sim}$. 
Again we made the empirical design choice of using 10 feature categories and 20 samples to cover most frequently encountered interesting driving behaviors through trial and error.

\paragraph{Iterative Refinement using Similarity Check}
Once {\em FeatureGPT} produces feature vectors for both the generated and original videos, we compare them to detect discrepancies. 
A large discrepancy in any feature indicates that the generated video is either missing a key behavior (negative gap) or introducing an unintended one (positive gap), and then we map the gap into natural language feedback (e.g., "there should be a leading vehicle overtaking behavior, please improve on that") and feed back into {\em ScriptGPT} to refine the SCENIC script. 
Gap thresholds $\tau_i$ for each feature $i$, as shown in Table \ref{table: pre-defined category}, are customized through empirical testing. We assign slightly larger thresholds to weather and road type features compared to other driving behavior features, as these are considered less critical for the testing objectives. This leniency allows for greater variety in environmental parameters, supporting the generation of diverse scenarios while ensuring that core driving behaviors remain accurately captured.

\begin{table}[htbp!]
\begin{center}
\caption{Pre-defined feature Category and Threshold}
\label{table: pre-defined category}
\begin{tabular}{ ||c|c|| } 
     \hline
     Pre-define Feature & Gap Threshold $\tau$ \\
     \hline
     Sunny / Rainy & $0.3$  \\ 
     Urban / Highway & $0.3$  \\ 
     Random Object on Road & $0.2$  \\ 
     Leading Vehicle Cruising & $ 0.2$  \\ 
     Leading Vehicle Stopped & $0.2$  \\ 
     Parallel Vehicle Cutting in & $0.2$  \\ 
     Parallel Vehicle Cruising & $0.2$  \\ 
     Parallel Vehicle Stopped & $0.2$  \\ 
     Behind Vehicle Overtaking & $0.2$ \\
     Opposite Vehicle Turning & $0.2$  \\ 
     \hline
\end{tabular}
\end{center}
\end{table}

\subsection{Case Study}
\label{sec:case-study}
\begin{figure}[H]
    \centering
    \includegraphics[width=\linewidth]{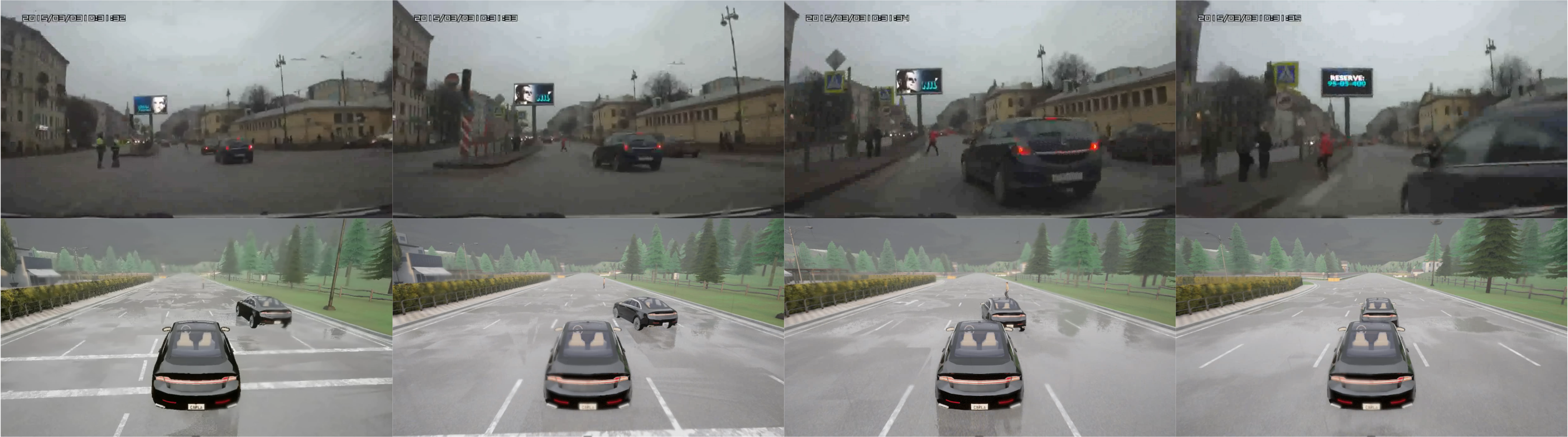}
    \caption{\small \textbf{Vehicle Cutting In with Pedestrian Crossing Scenario}: in the original dash camera video (top row), the vehicle on the right performs an emergency lane change to the left due to a jaywalking pedestrian in red. 
    In the generated scenario (bottom row) produced by our framework, the vehicle on the right exhibited a similar lane change behavior to the left to avoid a jaywalking pedestrian.}
    \label{fig:scenario-ped-crossing}
\end{figure}

\begin{figure}[htbp!]
    \centering
    \includegraphics[width=\linewidth]{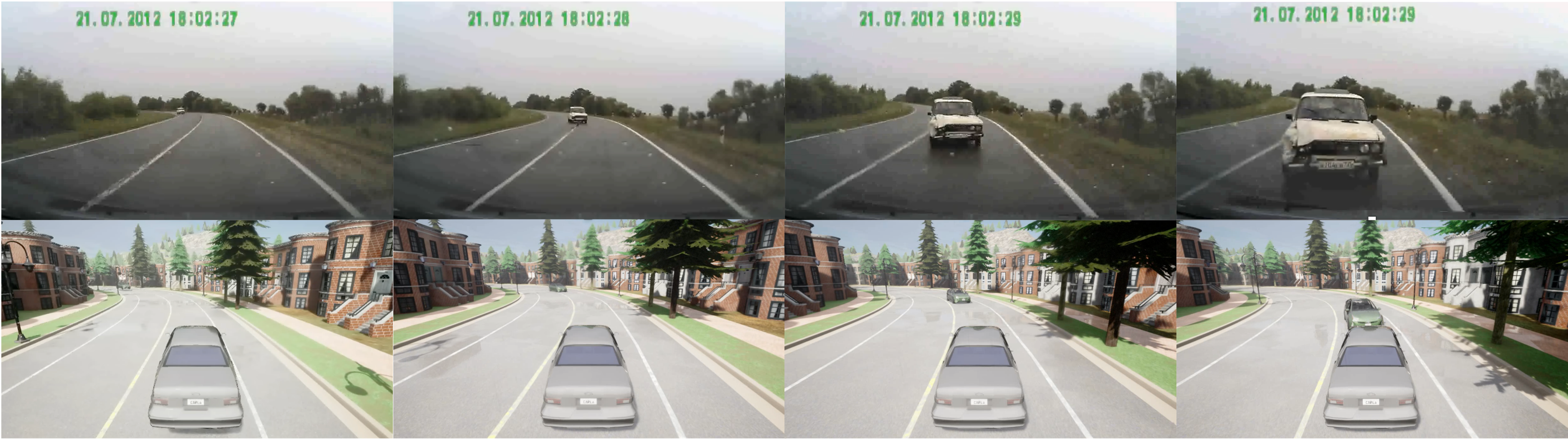}
    \caption{\small \textbf{Opposite Vehicle Invading Lane Scenario}: in the original dash camera video (top row), the vehicle on the opposite lane gradually swifts to ego's lane probably due to loss of focus. 
    In the generated scenario (bottom row) produced by our framework, the vehicle on the opposite lane exhibited a similar lane change behavior to switch to our lane and caused collision.}
    \label{fig:scenario-lane-invasion}
\end{figure}

\begin{figure}[htbp!]
    \centering
    \includegraphics[width=\linewidth]{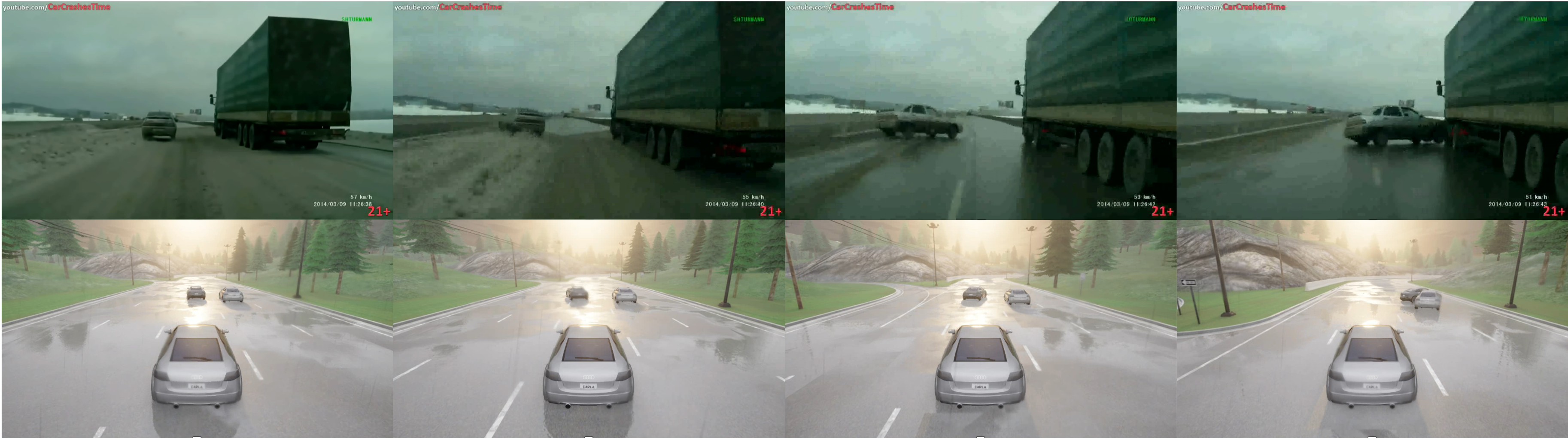}
    \caption{\small \textbf{Vehicle Spin Scenario}: in the original dash camera video (top row), the vehicle in front of the ego first spins to the left and then collided into the right vehicle. 
    In the generated scenario (bottom row) produced by our framework, the front vehicle exhibited a similar spin and collision behavior}
    \label{fig:scenario-spin}
\end{figure}

\begin{figure}[H]
    \centering
    \includegraphics[width=\linewidth]{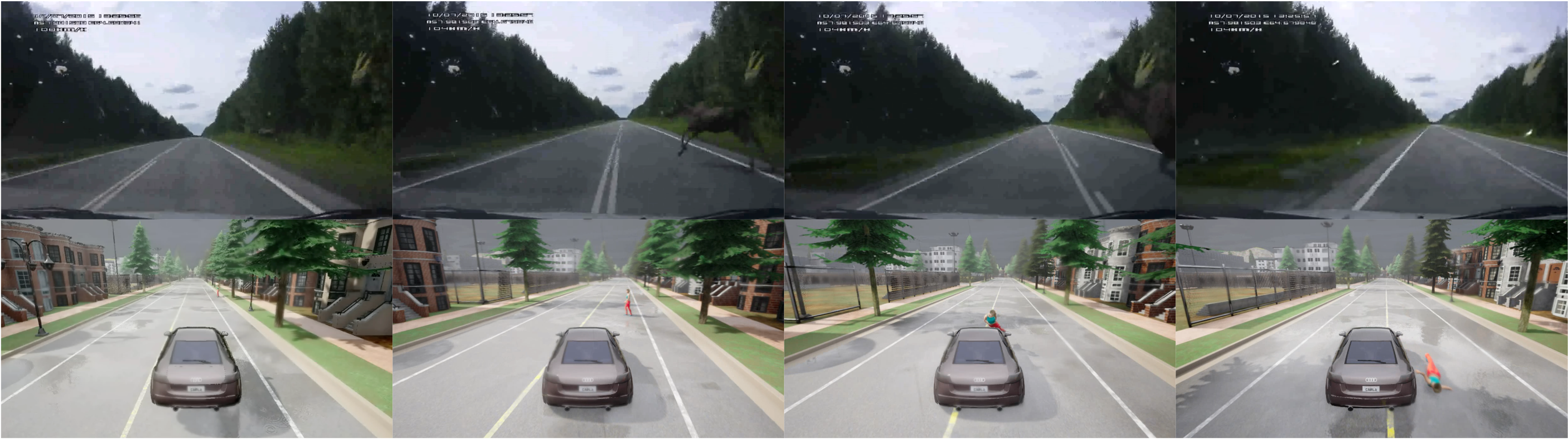}
    \caption{\small \textbf{Animal Crossing Scenario}: in the original dash camera video (top row), an animal attempted to cross the road, prompting the ego vehicle to perform an emergency lane change to the left. 
    In the generated scenario (bottom row) produced by our framework, the ego vehicle exhibited a similar lane change behavior to the left to avoid a jaywalking pedestrian (since CARLA does not have an animal model, the animal is replaced by a pedestrian).}
    \label{fig:scenario-animal-crossing}
\end{figure}

\begin{figure}[htbp!]
    \centering
    \includegraphics[width=\linewidth]{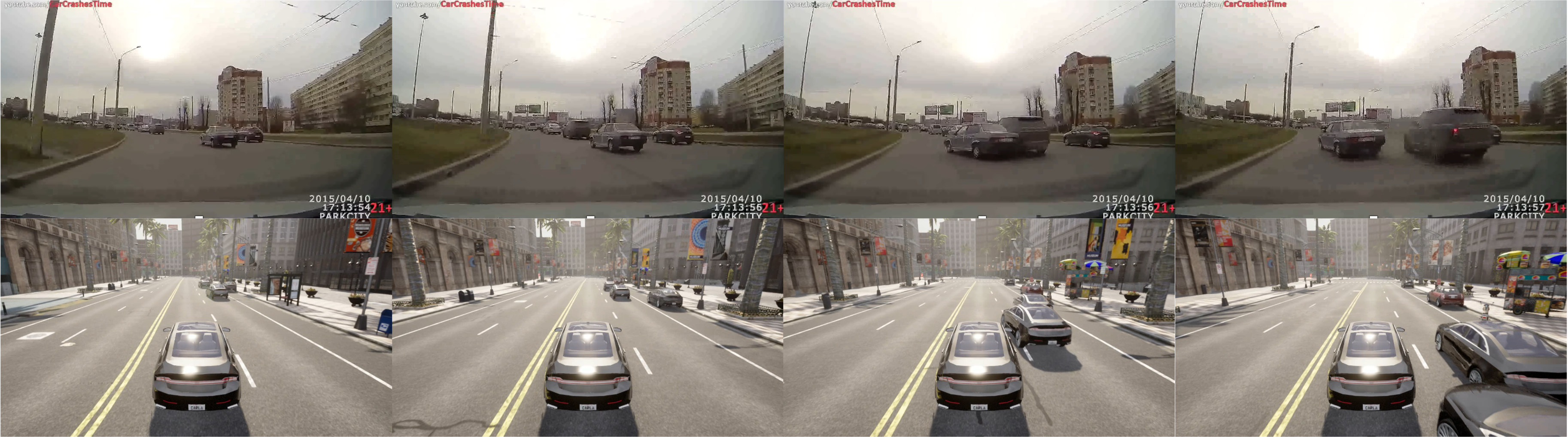}
    \caption{\small \textbf{Vehicle Cutting In with Stopped Object Scenario}: in the original dash camera video (top row), the vehicle on the right perform an emergency brake and tried lane change to the left due to a front parked vehicle. 
    In the generated scenario (bottom row) produced by our framework, the vehicle on the right exhibited a similar lane change behavior to the left to avoid the stopped vehicle and cause a collision. }
    \label{fig:scenario-lane-change}
\end{figure}

We present five interesting dashcam videos from the CCD dataset and generate their corresponding simulation scenarios using our framework, as shown in Figure \ref{fig:scenario-ped-crossing}, \ref{fig:scenario-lane-invasion}, \ref{fig:scenario-spin}, \ref{fig:scenario-animal-crossing}, \ref{fig:scenario-lane-change}. Importantly, each SCENIC script can correspond to multiple simulation scenarios. For example, a highway scenario may result in a 2-lane road or a 3-lane road, and a leading vehicle could be represented as either a truck or a sedan. This variability enables search-based testing by generating diverse scenarios that explore different environmental parameters while preserving the core driving behaviors.

In this section, we showcase one simulation scenario per SCENIC script due to page limitations. However, our framework inherently supports generating multiple testing scenarios from a single script, allowing systematic exploration of predefined features. As demonstrated in the figures, the generated scenarios faithfully capture the most essential driving behaviors observed in the dashcam videos while introducing controlled variations in other parameters, such as road configuration and vehicle types, to support robust ADS testing.

\subsection{Preliminary Qualitative Result}

\paragraph{Automated Pipeline} After completing the prompt engineering process, our framework is capable of automatically generating the 5 scenarios mentioned in Section \ref{sec:case-study} without any human intervention during the testing phase. 
We also extended our framework to evaluate 50 randomly selected accidents from the CCD dataset, and found that 32 out of 50 (64\%) scenarios generation can be fully automated without any human involvement, while 18 scenarios (36\%) encountered errors (i.e. generating a keyword or object that is not supported by SCENIC syntax), thus cannot ran automatically.

\paragraph{Time Efficiency} Our framework significantly reduces the time required for real-to-simulation scenario generation. For the 32 videos that can be automatically generated, it takes 1.5 minutes per scenario on average during the testing phase, including iterative refinement, to produce a SCENIC script of approximately 70 lines of code. In contrast, manually coding and debugging a similar real-to-simulation scenario could take an experienced engineer several hours. Even for the 36\% scenarios that cannot automatically finish, it only takes minutes for human to fix the syntax error to make the pipeline work again.

\paragraph{Advantage of Iterative Refinement} The 5 scenarios in Section \ref{sec:case-study} underwent 1-2 iterations of refinement, resulting in notable improvements in both accuracy and realism. In the extended evaluation of 50 accidents, iterative refinement happened in 17 scenarios (34\%), further showcasing its potential to enhance scenario quality and be generalized to more crash scenarios.

\paragraph{Framework Accuracy} While we have not yet conducted a formal quantitative analysis on the framework’s timing efficiency or objectively measure benefits of iterative refinement, the preliminary results provide strong evidence of the concept's validity. Our framework consistently captures the core driving behaviors from the original videos, indicating its effectiveness in generating accurate and realistic driving scenarios.

\section{Limitations \& Future Work}

\begin{figure}[htbp!]
    \centering
    \includegraphics[width=0.7\linewidth]{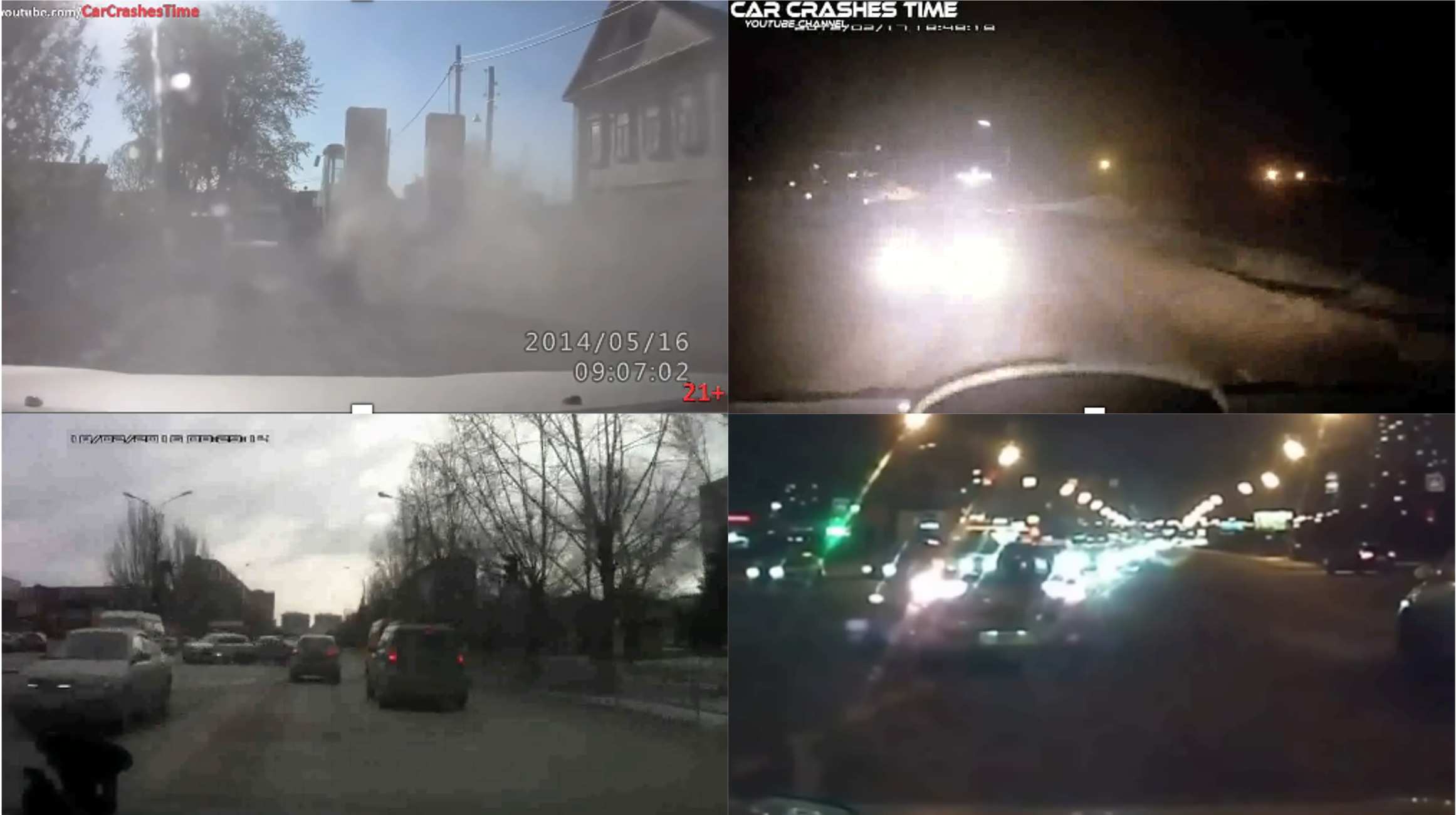}
    \caption{\small Scenarios from the Car Crash Dataset where our current framework cannot handle due to heavy traffic with multiple cars or compromised lighting environmental conditions at night}
    \label{fig:scenario-unrealizable}
\end{figure}

The current framework faces several challenges. 
It struggles with scenarios involving poor perception conditions or high complexity, as illustrated in Figure \ref{fig:scenario-unrealizable}. 
Additionally, the framework's performance may degrade when testing scenarios (e.g. multi-car complicated driving scenario at night like shown in Figure \ref{fig:scenario-unrealizable}) have not been encountered by {\em ScriptGPT} and {\em FeatureGPT} during the prompt engineering process.
Therefore, the performance of our framework heavily relies on carefully and manually selecting diverse "positive-examples" for prompt engineering.


For future work, we plan to conduct a human study to quantitatively assess the framework's time efficiency and accuracy. 
This will involve timing how long experts take to manually write real-to-simulation conversion scripts and asking them to rate the accuracy of our automated conversions. 
We also aim to improve the diversity of data samples used in the prompt engineering process to extend our framework's applicability across the entire Car Crash Dataset. 
Finally, we envision extending this video-to-video conversion framework to other domains, such as flying tasks or other robotics applications, broadening its use cases and potential impact.

\section{Conclusion}

In this paper, we have presented a novel framework for automatically converting real-world vehicle crash videos into simulation scenarios using prompt-engineered Video-Language Models. 
We have deploy multiple techniques including the similarity score vector metric and the iterative refinement process to ensure the generated scenarios closely align with the original videos. 
Despite the framework's current limitations, such as reliance on data diversity and the challenges of handling complex or unseen scenarios, through multiple examples, it demonstrates clear potential for improving ADS testing. 
Future work will focus on expanding data diversity, conducting quantitative human studies, and extending the framework to other robotics domains. 


\bigskip

\bibliography{references}    

\end{document}